\documentclass{article}
\usepackage{spconf,amsmath,graphicx}
\usepackage[ruled,vlined]{algorithm2e}
\usepackage[colorinlistoftodos]{todonotes}
\usepackage{caption}
\usepackage{subcaption}

\title{SIMPLE ONLINE AND REALTIME TRACKING WITH OCCLUSION HANDLING}

\twoauthors
{Mohammad Hossein Nasseri, Hadi Moradi, Reshad Hosseini}
{School of Electrical and Computer Engineering\\
University of Tehran\\
Tehran, Iran}
{Mohammadreza Babaee}
{Independent researcher\\
\\}
\begin{document}
\graphicspath{ {images/} }
%
\maketitle
\begin{abstract}
Multiple object tracking is a challenging problem in computer vision due to difficulty in dealing with motion prediction, occlusion handling, and object re-identification. Many recent algorithms use motion and appearance cues to overcome these challenges. But using appearance cues increases the computation cost notably and therefore the speed of the algorithm decreases significantly which makes them inappropriate for online applications. In contrast, there are algorithms that only use motion cues to increase speed, especially for online applications. But these algorithms cannot handle occlusions and re-identify lost objects. In this paper, a novel online multiple object tracking algorithm is presented that only uses geometric cues of objects to tackle the occlusion and re-identification challenges simultaneously. As a result, it decreases the identity switch and fragmentation metrics. Experimental results show that the proposed algorithm could decrease identity switch by 40\% and fragmentation by 28\% compared to the state of the art online tracking algorithms. The code is also publicly available\footnote{https://github.com/mhnasseri/sort\_oh}.
\end{abstract}
\section{Introduction}
\label{sec:intro}

Multiple Object Tracking (MOT) is a challenging problem in computer vision that has a wide variety of applications~\cite{babaee2017combined, babaee2016pixel}. In many applications such as autonomous driving, robot navigation, and visual surveillance real-time tracking is highly demanded. So, online MOT methods that only use current and previous frames can be used for these applications. Most recent MOT methods use tracking by detection pipeline which at first targets are detected by a detection algorithm and the results are passed to a data association algorithm to find trajectories ~\cite{weng2020gnn3dmot, bergmann2019tracking, papakis2020gcnnmatch}. Obviously, the quality of the detection algorithm significantly affects the result of the tracking algorithm. These algorithms utilize motion and appearance information in their data association part. However using the appearance feature imposes high computation cost and hence decreases the speed of algorithms. As a result, the maximum frame rate of the recent online MOT algorithms in MOT17\footnote{https://motchallenge.net/} dataset with a high MOTA metric is below $5$ Frame Per Second (FPS) even using high-performance hardware~\cite{weng2020gnn3dmot, papakis2020gcnnmatch, zhang2020multiplex}. On the other hand, algorithms that only use motion cues cannot overcome occlusion and object re-identification challenges. For example, the Simple Online and Real-time Tracking (SORT)~\cite{bewley2016simple} algorithm that only uses motion cues, achieves high FPS. However the number of ID switch and fragmentation metrics~\cite{milan2016mot16} are high that shows its low performance in occlusion handling and target re-identification.

In this paper, a new algorithm is proposed which only uses the location and size of detection bounding boxes to tackle occlusion without imposing high computation cost. Additionally, it can predict occlusions and re-identify lost targets. As a result, it decreases the ID switch and fragmentation metrics significantly. 

The rest of the paper is as follows: Section~\ref{sec:relatedwork} provides related work in the area of online multiple object tracking. Section~\ref{sec:approach} describes the proposed confidence based occlusion handling and target re-identification. In Section~\ref{sec:experiments}, the experiments are explained and the results are presented. Finally, in Section~\ref{sec:conclusion}, we draw our conclusion.

\begin{figure*}[!htb]
\centering
\includegraphics[width=18.2cm, height=1.2cm]{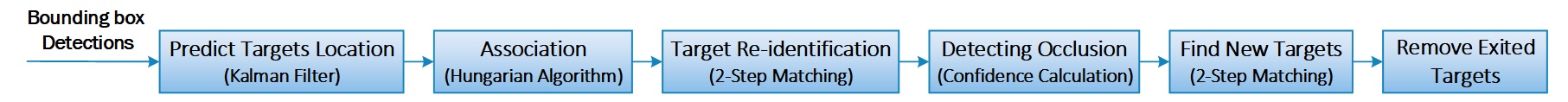}\\
\caption{The block diagram of the proposed algorithm}
\label{fig:block}
\end{figure*}

\section{related work}
\label{sec:relatedwork}

The majority of works in recent years use the tracking by detection paradigm. In this paradigm, first, a detection algorithm locates targets. For example, in MOT17 benchmark, DPM~\cite{felzenszwalb2008discriminatively}, FRCNN~\cite{ren2015faster}, and SDP~\cite{yang2016exploit} algorithms, which are based on Convolutional Neural Network (CNN), are used for target's bounding box detection. The second step is extracting features from bounding box detections, such as the motion and appearance features thet are widely used. For extracting motion features, the trajectory of each target should be predicted from the location and size of its bounding box. In some algorithms, filter-based methods such as Kalman filter~\cite{kalman1960new} with constant velocity assumption is used to predict the bounding box of targets~\cite{bewley2016simple,wojke2017simple}. In some other works, optical flow~\cite{wang2017online} and LSTM~\cite{milan2016online} are employed. In~\cite{leal2011everybody}, the motion of targets in crowded scenarios are predicted by utilizing more complex methods such as the social force model. Recently reinforcement learning is proposed for the prediction of target location in successive frames~\cite{chen2018real}. 

For appearance features, the early works used handcrafted features such as color histograms~\cite{zhang2008global}. Recent works learn features from CNN networks including Siamese network~\cite{kim2016similarity}, correlation filters~\cite{kim2017multi} and auto-encoders~\cite{feng2017using}. However, the computation cost of CNN-based methods is high that gives rise to a significant decrease in the speed of algorithms, even on high-performance hardware. Concequently, there are hardly suitable for online applications.

The third step is calculating the similarity between detections and targets. For calculating affinity from predicted motion, the location and size of bounding boxes are considered and metrics such as Intersection over Union (IoU) is calculated~\cite{bewley2016simple,wojke2017simple}. Also for obtaining similarity between object pairs based on an appearance feature, cosine similarity~\cite{weng2020gnn3dmot} and other CNN based methods~\cite{babaee2019dual} as well as LSTM variants~\cite{kim2018multi} are used. 

The final step is associating detections to targets. For this task different approaches are used such as Hungarian algorithm~\cite{bewley2016simple,wojke2017simple}, dynamic programming~\cite{ullah2018directed}, and multiple hypothesis tracking~\cite{kim2015multiple}. Recently reinforcement learning~\cite{ren2018collaborative} and graph-based methods~\cite{liugsm} are employed for the association task. Alongside the tracking by detection paradigm, some works combine detection and tracking tasks. For example, in~\cite{bergmann2019tracking}, a regressor based detector is used for tracking objects. Besides, There are end-to-end methods for object tracking that perform both detection and tracking at the same time~\cite{wangjoint, zhou2020tracking}.

The most referenced algorithms for online object tracking are SORT~\cite{bewley2016simple} and Deep SORT\cite{wojke2017simple}. In the Deep SORT algorithm, the appearance cue is added to the SORT algorithm to improve its weakness in target re-identification and imposes higher computation cost. This makes the Deep SORT algorithm inappropriate for real-time applications and decreases the speed of the algorithm dramatically. In this paper, an algorithm based on the SORT algorithm is proposed that deals with occlusion handling and target re-identification efficiently without using appearance cues and losing the speed of the algorithm. In the proposed algorithm, occlusion handling and target re-identification are tackled using only the location and size of bounding boxes and a comparable result to the Deep Sort algorithm is achieved at a very lower computation cost.

\section{Approach}
\label{sec:approach}

The proposed algorithm is inspired from ~\cite{bewley2016simple}, which is based on tracking-by-detection paradigm. The main steps of the proposed algorithm is shown in Fig.~\ref{fig:block}. The detected bounding boxes are given as input to the algorithm. These bounding boxes at frame $t$ are denoted as $\mathbf{D_t}=\{D^1_t, D^2_t, ..., D^{K_t}_t\}$ where $K_t$ is the number of detections at frame $t$. Also the targets at frame $t-1$ are indicated as $\mathbf{T_{t-1}}=\{T^1_{t-1}, T^2_{t-1}, ...,$ $ T^{P_{t-1}}_{t-1}\}$ where $P_{t-1}$ is the number of targets at frame $t-1$. The prediction of these targets at current frame are denoted as $\mathbf{\hat{T}_t}$ which is another input to the association step of the proposed algorithm. After the association step, most of detections are matched with targets and considered as new observation for correcting estimation and a few of them are not matched. The matched detections at frame $t$ are indicated by $\mathbf{\tilde{D}_t}$ and unmatched detections are depicted by $\mathbf{\bar{D}_t}$. After the association step, some targets are matched with detections, $\mathbf{\tilde{T}_t}$, a few of them are marked as occluded, that are $\mathbf{O_t}$ and a few of them are remained unmateched, $\mathbf{\bar{T}_t}$. At the end, some new targets could be created and added to $\mathbf{\tilde{T}_t}$ or some targets may be discarded and removed from $\mathbf{\bar{T}_t}$.

The first step of the algorithm is to model objects and their motions to predict their location in future frames. Like~\cite{bewley2016simple}, the state of targets are modeled as:
\begin{equation}
x=[u, v, s, r, \dot{u}, \dot{v}, \dot{s}]
\end{equation}
In the above equation, $u$ and $v$ represent the center, $s$ represents the area, and $r$ represents the aspect ratio of a bounding box. The $\dot{u}, \dot{v}$ and $\dot{s}$ are the rate of change for the corresponding variables. A Kalman filter with a constant velocity model is utilized to model the motion of objects. This filter predicts the location and size of each target in the next frames.

For associating detections to targets, the Intersection over Union (IoU) is calculated. This metric is equal to the intersection of the bounding box of detection with a bounding box of each target divided by the union area of these two bounding boxes:
\begin{equation}
IoU = \frac{I(bb_D, bb_T)}{A(bb_D)+A(bb_T)-I(bb_D, bb_T)}
\end{equation}
in which $bb_D$ and $bb_T$ stand for the bounding box of a detection and the bounding box of a target, respectively. Also, $A$ represents the area of a bounding box and $I$ represents the intersection area of two bounding boxes. In the association step, the Hungarian algorithm uses IoU values as a similarity metric to assign detection boxes to the targets.

\subsection{Target's Confidence}

The main idea of the proposed algorithm is introducing a confidence value for each target during tracking. Basically, in online tracking algorithms, the targets matched with detections are marked as live targets and the rest are kept as reserved targets or are discarded. In the proposed algorithm, a confidence measure is defined that plays a key role in determining if a target should be marked as occluded. This confidence measure is a representation of the temporal and spatial characteristics of a target.  In the case of temporal characteristics, if a target is observed for several successive frames, the probability of its existence increases. In contrast, if the target is not detected for several consecutive frames, its certainty decreases. In the case of spatial characteristics of a target, if the bounding box of a target is larger, it means that the target covers larger area of the image and the confidence in such a target is higher. It may be noted that for targets with similar sizes, the bigger bounding box is correlated to being closer to the camera. Having discussed all relevant factors, the confidence of $i^{th}$ target ($C_i$) is defined as:
\begin{equation}
C_i = min(1, \alpha * \frac{Age_{i}}{t_{so}} * \frac{A_i}{A_{avg}}), 
\label{eq:conf}
\end{equation}
in which $\alpha$ is a hyperparameter. $Age_{i}$ is the number of frames a target existed since its appearance. The $t_{so}$ or time since observed, is the number of successive frames that a target is not matched with any detection. $A_i$ denotes the area of the target and $A_{avg}$ is the average area of all targets. $A_{avg}$ is used to normalize the area of the bounding box of the target, to eliminate the effects of the camera position.

\subsection{Occlusion Handling}

In reality, a target may be partially or fully occluded by objects or other targets. The detection algorithm, only detects targets, in our case persons. So, there are not any corresponding detections for other objects. To detect occlusion by objects, the confidence metric is used. When a target does not match with any detection, if its confidence is greater than the object confidence threshold, which is represented as $C_{O}$, the target is marked as occluded. This can cover the weakness of the detection algorithm when there is no corresponding detection for a target due to an error in the detection algorithm.

The second kind of occlusion is when some other targets lie between the camera and a target. So, there is no corresponding detection to that covered target. In such cases, the intersection of the estimated bounding box of that target with the bounding box of the covering target is high. This property can help to detect such a kind of occlusion. The IoU metric is not appropriate for detecting occlusion. When the area of a target is very larger than another target, the IoU decreases since the intersection is divided into the total area of targets including the bigger one. This situation is common in occlusion, because the target which is closer to the camera and is seen bigger, may occlude a far target which is seen smaller. To overcome this weakness, a novel metric, the so-called Coverd Percent (CP) is proposed, that measure how many percent of a target is covered by another target. The CP for the $i^{th}$ target is defined as:

\begin{equation}
CP_{i} = \frac{I(bb_{i}, bb_{j})}{A(bb_{i})}
\end{equation}

To detect target-target occlusion, minimum thresholds for CP and confidence metrics are defined. They are represented as $CP_{min}$ and $C_{T}$, respectively. If the covered percent of an unmatched target with any of other targets is bigger than $CP_{min}$ and its confidence is larger than $C_{T}$ at the same time, it is marked as occluded, too.

\begin{figure}[t]
\begin{minipage}[b]{0.48\linewidth}
  \centering
  \centerline{\includegraphics[height=5cm]{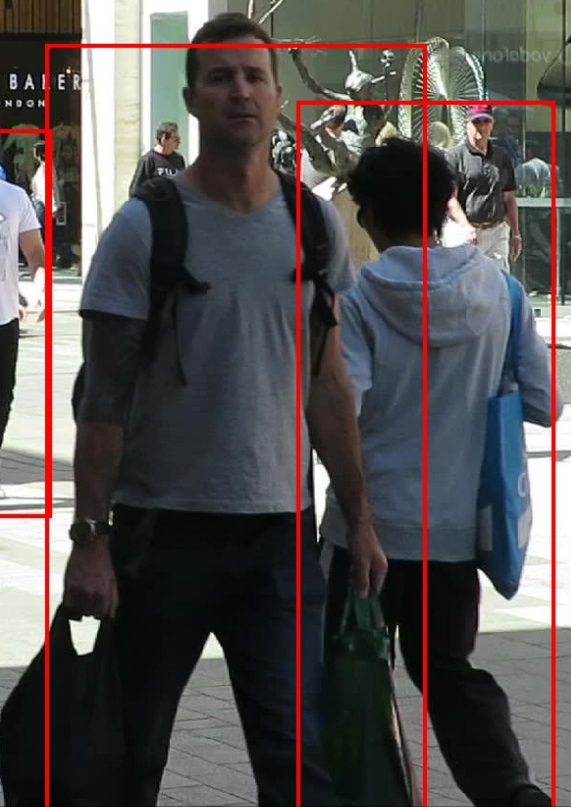}}
  \centerline{(a)}
\end{minipage}
\begin{minipage}[b]{0.48\linewidth}
  \centering
  \centerline{\includegraphics[ height=5cm]{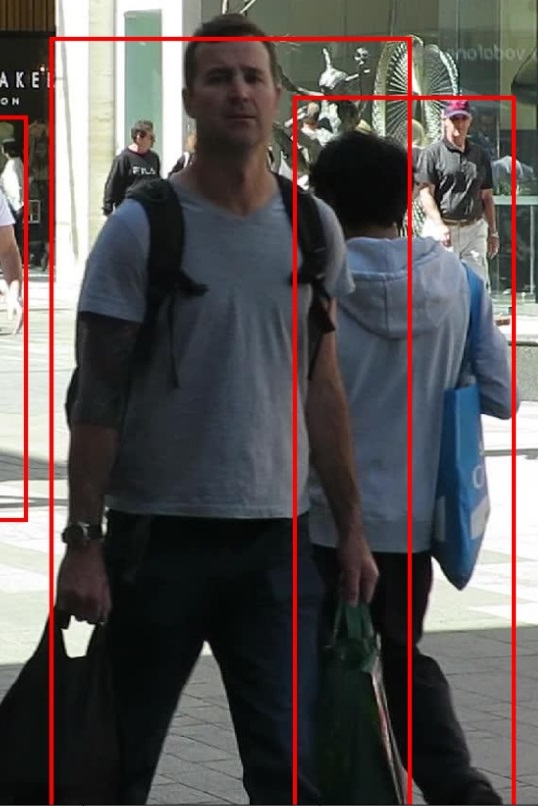}}
  \centerline{(b)}
\end{minipage}
\caption{Gradual occlusion. The detections are depicted by red rectangles. (a) A target is going to be occluded behind another target. (b) A part of target is occluded which leads to size reduction of its bounding box.}
\label{fig:grad}
\end{figure}

For occluded targets, there are no corresponding detections to correct their estimation in the update step of their Kalman filters. For these targets, only the rate of area's change is halved to prevent the size of their estimated bounding boxes decrease unrealistically. This is because occlusion does not happen at once. First, part of the target becomes occluded and gradually becomes fully occluded. So, the size of the corresponding detection is reduced in frames leading to the occlusion, because the bounding box surrounds the observable part of the target. 

\begin{equation}
\dot{s}_{t+1} = \frac{\dot{s}_t}{2}
\end{equation}
The gradual occlusion of targets and the decrease of the size of the corresponding detection is shown in Fig.~\ref{fig:grad}.

\begin{figure*}[htb]
\centering
\begin{tabular}{cccc}
\includegraphics[width=3.05cm, height=5cm]{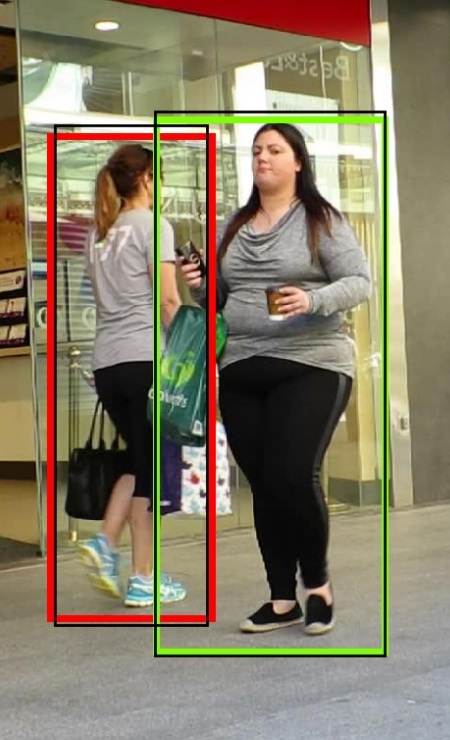}&
\includegraphics[width=3.05cm, height=5cm]{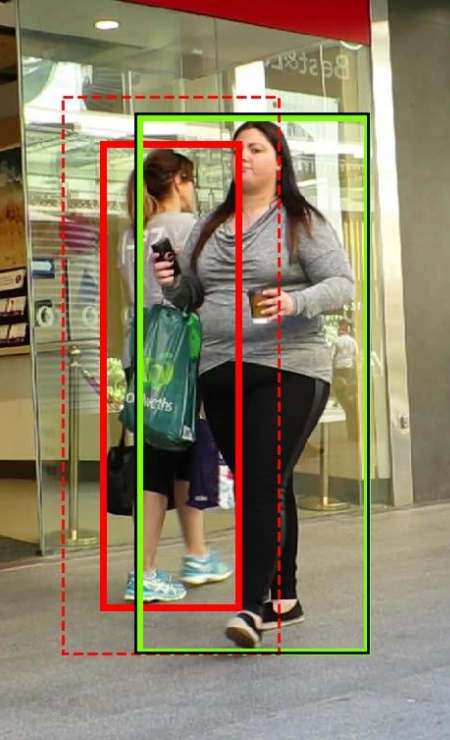}&
\includegraphics[width=3.05cm, height=5cm]{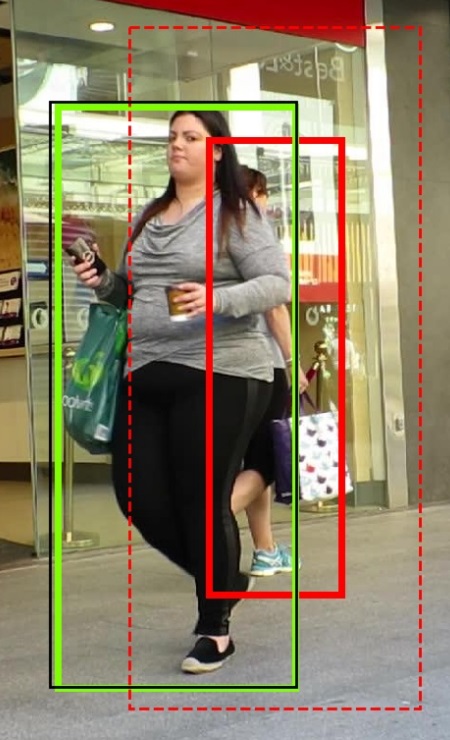}&
\includegraphics[width=3.05cm, height=5cm]{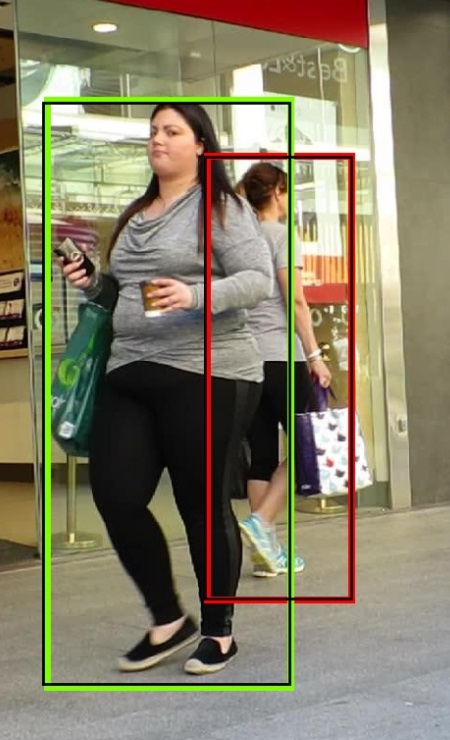}\\
(a)& (b)& (c)& (d)\\
\end{tabular}
\caption{Example of re-identifying an occluded target. Detections are depicted with thin black rectangles, Targets are depicted with colored thick bounding boxes, extended bounding boxes are depicted with dashed rectangles. (a) A target is going to be occluded behind another target. (b) A target is occluded by another target and an extended bounding box is assigned to it. (c) The target is occluded behind another target for several frames and the size of its extended bounding box is increased. (d) The extended bounding box helps to re-identify the occluded target.}
\label{fig:oh}
\end{figure*}

\subsection{Target Re-identification}

Target re-identification is done in a cascade matching manner. In the first step, the IoU between detections and all existing targets, including occluded targets are calculated and the association is performed using the Hungarian algorithm. Because the estimated bounding boxes of occluded targets do not correct in the update step of the Kalman filter, there may become a considerable difference between their estimation and real location. As a result, they may not match in the first step. 

In the second step, to help the re-identification of unmatched occluded targets, their bounding boxes are extended according to their uncertainty. This uncertainty increases in every next frame where the target is not observed. Then the IoU between extended bounding boxes and remaining unmatched detections should be calculated. Because the sizes of occluded targets do not extend in reality and only their uncertainty increase, the extended IoU is calculated as below:
\begin{equation}
IoU_{ext} = \frac{I(bb_D, bb_{ext_T})}{A(bb_D)+A(bb_T)-I(bb_D, bb_{ext_T})}
\end{equation}
In which $bb_{ext_T}$ is the extended bounding box of the occluded target. The extended bounding box is only used for computing intersection and the area of the estimated bounding box is used in the denominator. To complete the second step, the calculated $IoU_{ext}$ is passed to the Hungarian algorithm for the association. In Fig.~\ref{fig:oh}, the extended bounding box of an occluded target is depicted by dashed lines.

\subsection{Creating New Targets}

\begin{figure}[hb]
\begin{minipage}[b]{0.3\linewidth}
  \centering
  \centerline{\includegraphics[width=1.7cm]{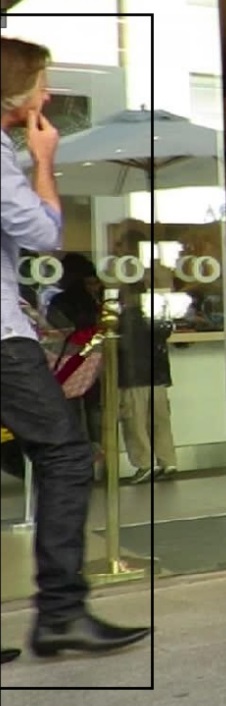}}
  \centerline{(a)}
\end{minipage}
\begin{minipage}[b]{0.3\linewidth}
  \centering
  \centerline{\includegraphics[width=1.7cm]{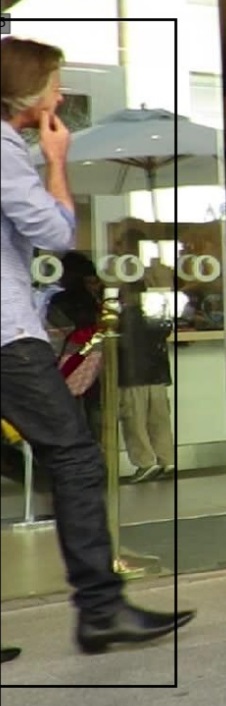}}
  \centerline{(b)}
\end{minipage}
\begin{minipage}[b]{0.3\linewidth}
  \centering
  \centerline{\includegraphics[width=1.8cm]{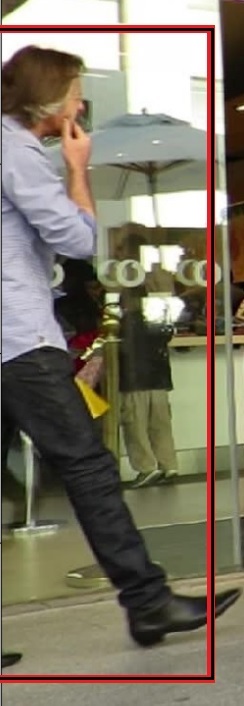}}
  \centerline{(c)}
\end{minipage}
\caption{New target creation after three frames. The detections are shown with thin black rectangles and the target is shown with red thick rectangle. (a) A new target is seen for the first time. (b) The new target is seen for the second time. (c) The new target is seen for the third time and a target is associated with it.}
\label{fig:newtarget}
\end{figure}

During tracking, new targets may be entered into the scene. In the first few frames, all unmatched detections, represented by $\mathbf{\bar{D}_t}$ are considered as new targets. But after that, a new target is created after the probability of its existence increases. This is achieved when unmatched detections are presented near a location for three successive frames. For detecting this situation, the unmatched detection of the current frame, the previous frame, and the two frames before are used. To assign unmatched detections from different frames to each other, a process similar to associating detections to targets is performed. At first, an IoU matrix between $\bar{D}_t$ and $\bar{D}_{t-1}$ and another IoU matrix between $\bar{D}_{t-1}$ and $\bar{D}_{t-2}$ are calculated. Then each calculated IoU matrix is passed to the Hungarian algorithm. If three unmatched detections from different frames are assigned to each other, a new target is created and its Kalman filter is initialized using corresponding unmatched detections.
The process of creating a new target is shown in Fig.~\ref{fig:newtarget}.

\begin{table*}[th!]
\centering
\begin{tabular}{l c c c c c c c c c c}
\hline
& & \textbf{MOTA} $\uparrow$ & \textbf{MOTP} $\uparrow$ & \textbf{MT} $\uparrow$ & \textbf{ML} $\downarrow$ & \textbf{IDS} $\downarrow$ & \textbf{FM} $\downarrow$ & \textbf{FP} $\downarrow$ & \textbf{FN} $\downarrow$ & \textbf{FPS} $\uparrow$\\
\hline
KDNT~\cite{yu2016poi} & BATCH & \textcolor{blue}{68.2} & 79.4 & \textcolor{blue}{41.0\%} & \textbf{19.0\%} & 933 & 1093 & 11479 & \textcolor{blue}{45605} & 0.7 \\
LMP\_p~\cite{keuper2016multi} & BATCH & \textbf{71.0} & \textbf{80.2} & \textbf{46.9\%} & \textcolor{blue}{21.9\%} & \textcolor{blue}{434} & \textbf{587} & \textcolor{blue}{7880} & \textbf{44564} & 0.5 \\
MCMOT\_HDM~\cite{lee2016multi} & BATCH & 62.4 & 78.3 & 31.5\% & 24.2\% & 1394 & 1318 & 9855 & 57257 & \textbf{35} \\
NOMTwSDO16~\cite{choi2015near} & BATCH & 62.2 & \textcolor{blue}{79.6} & 32.5\% & 31.1\% & \textbf{406} & \textcolor{blue}{642} & \textbf{5119} & 63352 & \textcolor{blue}{3} \\
\hline
EAMTT~\cite{sanchez2016online} & ONLINE & 52.5 & 78.8 & 19\% & 34.9\% & 910 & \textbf{1321} & \textbf{4407} & 81223 & 12 \\
POI~\cite{yu2016poi} & ONLINE & \textbf{66.1} & \textcolor{blue}{79.5} & \textbf{34\%} & \textcolor{blue}{20.8\%} & \textcolor{blue}{805} & 3093 & \textcolor{blue}{5061} & \textbf{55914} & 10 \\
SORT~\cite{bewley2016simple} & ONLINE & 59.8 & \textbf{79.6} & 25.4\% & 22.7\% & 1423 & 1835 & 8698 & 63245 & \textcolor{blue}{60} \\
Deep SORT~\cite{wojke2017simple} & ONLINE & \textcolor{blue}{61.4} & 79.1 & \textcolor{blue}{32.8\%} & \textbf{18.2\%} & \textbf{781} & 2008 & 12852 & \textcolor{blue}{56668} & 40 \\
\textbf{Proposed} & ONLINE & 61.1 & 79.04 & 31.62\%& 21.34\% & 848 & \textcolor{blue}{1331} & 12296 & 57738 & \textbf{162.7} \\
\hline
\end{tabular}
\caption{The results of our proposed tracking algorithm compared to the results of state-of-the-arts on MOT16. The best are in the bold format. The second best are in the blue format.}
\label{tab:res_mot16}
\end{table*}

\begin{table*}[th!]
\centering
\begin{tabular}{l c c c c c c c c c}
\hline
& \textbf{Detection} & \textbf{MOTA} $\uparrow$ & \textbf{MT} $\uparrow$ & \textbf{ML} $\downarrow$ & \textbf{IDS} $\downarrow$ & \textbf{FM} $\downarrow$ & \textbf{FP} $\downarrow$ & \textbf{FN} $\downarrow$ & \textbf{FPS} $\uparrow$\\
\hline
Tractor & ALL & \textcolor{blue}{53.5} & \textcolor{blue}{19.5\%} & \textcolor{blue}{36.6\%} & \textcolor{blue}{2072} & \textcolor{blue}{4611} & \textbf{12201} & \textcolor{blue}{248047} & \textcolor{blue}{1.5}\\
GCNNMatch & ALL & \textbf{57.0} & \textbf{23.3\%} & \textbf{34.6\%} & \textbf{1957} & \textbf{2798} & \textcolor{blue}{12283} & \textbf{228242} & 1.3\\
\textbf{Proposed} & ALL & 44.3 & 14.4\% & 45.6\% & 2191 & 5243 & 21796 & 290065 & \textbf{137.5}\\
\hline
Tractor & DPM & \textcolor{blue}{52.2} & \textcolor{blue}{14.9\%} & \textbf{37.5\%} & 635 & - & \textbf{2908} & \textcolor{blue}{86275} & -\\
GCNNMatch & DPM & \textbf{55.5} & \textbf{21.5\%} & \textcolor{blue}{37.6\%} & \textcolor{blue}{564} & \textbf{782} & \textcolor{blue}{2937} & \textbf{80242} & -\\
\textbf{Proposed} & DPM & 29.7 & 5.5\% & 64.7\% & \textbf{530} & \textcolor{blue}{1448} & 3048 & 128696 & -\\
\hline
Tractor & FRCNN & \textcolor{blue}{52.9} & \textcolor{blue}{16.2\%} & \textcolor{blue}{34.7\%} & \textcolor{blue}{648} & - & \textbf{3918} & \textcolor{blue}{83904} & -\\
GCNNMatch & FRCNN & \textbf{56.1} & \textbf{22.3\%} & \textbf{33.9\%} & \textbf{647} & \textbf{934} & \textcolor{blue}{4015} & \textbf{77950} & -\\
\textbf{Proposed} & FRCNN & 44.9 & 14.7\% & 40.3\% & 795 & \textcolor{blue}{1571} & 9102 & 93669 & -\\
\hline
Tractor & SDP & 55.3 & 18.1\% & 32.9\% & \textcolor{blue}{789} & - & \textcolor{blue}{5375} & 77868 & -\\
GCNNMatch & SDP & \textbf{59.5} & \textbf{26.0\%} & \textcolor{blue}{32.4\%} & \textbf{746} & \textbf{1082} & \textbf{5331} & \textcolor{blue}{70050} & -\\
\textbf{Proposed} & SDP & \textcolor{blue}{58.4} & \textcolor{blue}{22.8\%} & \textbf{32.0\%} & 866 & \textcolor{blue}{2224} & 9646 & \textbf{67700} & -\\
\hline
\end{tabular}
\caption{The results of our proposed tracking algorithm compared to the results of state-of-the-art algorithms on MOT17. The best are in the bold format. The second best are in the blue format.}
\label{tab:res_mot17}
\end{table*}

\subsection{Removing Targets}

An unmatched target is removed, when its uncertainty increases above a threshold. Like the occlusion handling section, the uncertainty is proportional to the number of frames in which the target is visible. Any unmatched target is retained for at least $k_{min}$ frames. But more confident targets are retained for more frames proportional to their age. To prevent retaining targets for more than needed, targets are retained for at most $k_{max}$ frames. An unmatched target is retained when:
\begin{equation}
t_{su} > min(k_{min}+\frac{Age_{T}}{c_{k}}, k_{max})
\end{equation}
in which $t_{su}$, time since updated, is the number of successive frames the estimated bounding box of a target is not corrected. Because Kalman filters of occluded targets are updated, they do not enter the deletion process.
\begin{algorithm}[!ht]
\caption{SORT with occlusion handling}
\DontPrintSemicolon
\LinesNumbered
\SetKwProg{Fn}{Function}{}
\KwData{\textbf{Data:} $D_{t}$, $T_{t-1}$}\;
\KwResult{$\bar{D}_t, T_{t}, \bar{T}_t$}
$\hat{T}_t = KF\_Predict(T_{t-1})$\;
$A_{avg} = Area\_Average(\hat{T}_t)$\;
$\tilde{T}_t, \tilde{D}_t, O_t, \bar{T}_t, \bar{D}_t = Associate(\hat{T}_t, D_t, A_{avg})$\;
$\tilde{T}_t=KF\_Correction(\tilde{T}_t, \tilde{D}_t)$\;
$O_t = Correction(O_t)$\;
$T_{t} \leftarrow \tilde{T}_t + \bar{T}_t + O_t$\;
\;
\Fn{Associate($\hat{T}_t, D_{t}, A_{avg}$)}{
\textbf{Cascade Matching:}\;
$P = IoU(D_{t}, \hat{T}_t)$\;
$\tilde{T}_t, \tilde{D}_t, \bar{T}_t, \bar{D}_t = Hungarian(P)$\;
\textbf{Target Re-identification:}\;
$P_{d} = IoU(\bar{D}_t, \bar{D}_{t-1})$\;
$P_{ext} = IoU_{ext}(\bar{T}_t, \bar{D}_t)$\;
$\tilde{T}^n_t, \tilde{D}^n_t = 2\_Step\_Matching(P_{d}, P_{ext})$\;
$\tilde{T}_t \leftarrow \tilde{T}_t + \tilde{T}^n_t$\;
$\bar{T}_t \leftarrow \bar{T}_t - \tilde{T}^n_t$\;
$\tilde{D}_t \leftarrow \tilde{D}_t + \tilde{D}^n_t$\;
$\bar{D}_t \leftarrow \bar{D}_t - \tilde{D}^n_t$\;
\textbf{Detecting Occlusion:}\;
$O_t \leftarrow \emptyset$\;
$P = CP(\hat{T}_t)$\;
\For{$ u \in \bar{T}_t$}{
$C_u = min(1,\alpha*\frac{Age_{u}}{t_{so_{u}}}*\frac{A_{u}}{A_{avg}})$\;
$P_{u} = max(P)$\;
\uIf{$C_u > C_O$}{
$O_t \leftarrow O_t + u$\;
$\bar{T}_t \leftarrow \bar{T}_t - u$\;
} 
\uElseIf{$C_u > C_T$ \textbf{and} $P_{u}>min\_coverage$}{
$O_t \leftarrow O_t + u$\;
$\bar{T}_t \leftarrow \bar{T}_t - u$\;
} 
} 
} 
\label{alg:sort_oh}
\end{algorithm}

\begin{algorithm}[!ht]
\caption{Target creation and removal}
\DontPrintSemicolon
\LinesNumbered
\SetKwProg{Fn}{Function}{}
\KwData{\textbf{Data:} $\bar{D}_t, \bar{D}_{t-1}, \bar{D}_{t-2}, T_t, \bar{T}_t$}\;
\KwResult{$\bar{D}_t, \bar{D}_{t-1}, T_t$}

$N_t \leftarrow \emptyset$\;
$R_t \leftarrow \emptyset$\;
\If{frame\_num $<$ min\_hits}{
$N_t \leftarrow \bar{D}_t$\;
$\bar{D}_t \leftarrow \emptyset$\;
} 
\Else{
$N_t = Find\_New\_Target(\bar{D}_t, \bar{D}_{t-1}, \bar{D}_{t-2})$\;
} 
$R_t = Remove\_Target(\bar{T}_t)$\;
$T_{t} \leftarrow T_t + N_t - R_t$\;
\;

\Fn{Find\_New\_Target($\bar{D}_t, \bar{D}_{t-1}, \bar{D}_{t-2}$)}{
$N_t \leftarrow \emptyset$\;
$P_{d} = IoU(\bar{D}_t, \bar{D}_{t-1})$\;
$P_{db} = IoU(\bar{D}_{t-1}, \bar{D}_{t-2})$\;
$N_t, N_{t-1} = 2\_Step\_Matching(P_{d}, P_{db})$\;
$\bar{D}_t \leftarrow \bar{D}_t - N_t$\;
$\bar{D}_{t-1} \leftarrow \bar{D}_{t-1} - N_{t-1}$\;
} 
\;
\Fn{Remove\_Target($\bar{T}_t$)}{
$R_t \leftarrow \emptyset$\;
\For{$ u \in \bar{T}_t$}{
\If{$t_{su_{u}} > min(k_{min}+\frac{Age_{u}}{c_{k}}, k_{max})$}{
$R_{t} \leftarrow R_{t} + u$\;
} 
} 
} 
\label{alg:target_oh}
\end{algorithm}

\section{Experiments}
\label{sec:experiments}

To compare the proposed algorithm with SORT~\cite{bewley2016simple} and DEEP SORT~\cite{wojke2017simple} algorithms, it is evaluated on MOT16 benchmark using private detections from POI~\cite{yu2016poi} paper. For a fair comparison, like POI and DEEP SORT papers, detections have been thresholded at a confidence score of 0.3. Also for comparing the proposed algorithm with the most recent algorithms, their results on MOT17 are presented.

\subsection{Evaluation Metrics}

For evaluating multiple object tracking algorithms, the frequently used CLEAR MOT metrics~\cite{bernardin2008evaluating} are reported, including Multiple Object Tracking Accuracy (MOTA), and Multiple Object Tracking Precision (MOTP). Also, other popular metrics are used including Mostly Tracked (MT), Mostly Lost (ML), the number of False Negatives (FN), False Positives (FP), ID-Switches (IDS), and track Fragmentation (FM).

\subsection{Results}


The result of running the proposed algorithm on the MOT16 benchmark alongside the result of baseline algorithms such as SORT and DEEP SORT come in the table ~\ref{tab:res_mot16}. In the MOTA and MOTP metrics, the results of the proposed algorithm are comparable with other online algorithms and it is slightly lower than the Deep SORT algorithm. For the IDS metric, in comparison to the SORT algorithm, the value from 1423 is decreased to 848 which is a 40\% reduction in this metric and it is only 8\% above the Deep SORT algorithm. Also, in fragmentation metric, the score of the proposed algorithm is 1331 which is 28\% lower than the SORT score which is 1835, and is even 34\% lower than the Deep SORT algorithm with the value of 2008. This is because the Deep SORT algorithm does not detect occlusions and uses only the appearance to re-identify targets after they appear again. All of these comparable or even better results are achieved with the only use of geometric features and with a much lower computation cost that leads to a much higher speed, even with ordinary hardware.

The MOT17 benchmark includes three sets of detections for every sequence. The detection algorithms are DPM, FRCNN, and SDP. The results of a tracking algorithm on this benchmark are the aggregation of its results in each detection set. In Table~\ref{tab:res_mot17} the result of two state-of-the-art tracking algorithms and the proposed algorithm are separated for each detection set.

Between these three detection algorithms, the DPM has the lowest performance. The performance of the FRCNN algorithm is better than DPM, but lower than SDP and the SDP algorithm has the best performance. For the DPM and FRCNN detection sets, there is a significant difference between the result of the proposed algorithm and these two algorithms. But for the SDP detection set, the results of the proposed algorithm are better than results of the algorithm proposed in~\cite{bergmann2019tracking} and is comparable with results of algorithm discussed in~\cite{papakis2020gcnnmatch}. These two tracking algorithms use appearance features and their speed even with high-performance hardware is lower than 2 FPS which is not appropriate for real-time applications. But the proposed algorithm reaches comparable results with the only use of geometric features while it is much faster than them.

\begin{figure*}[htb]
\centering
\begin{tabular}{ccc}
\includegraphics[width=5cm, height=3.75cm]{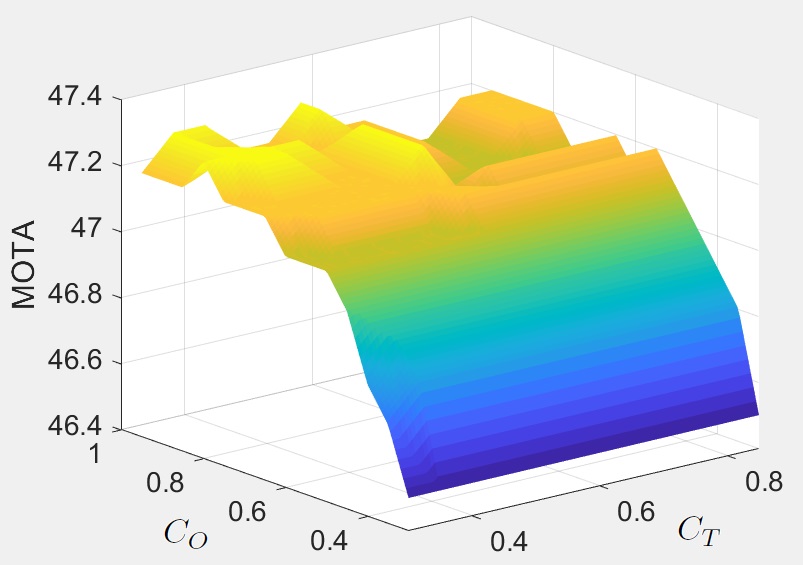}&
\includegraphics[width=5cm, height=3.75cm]{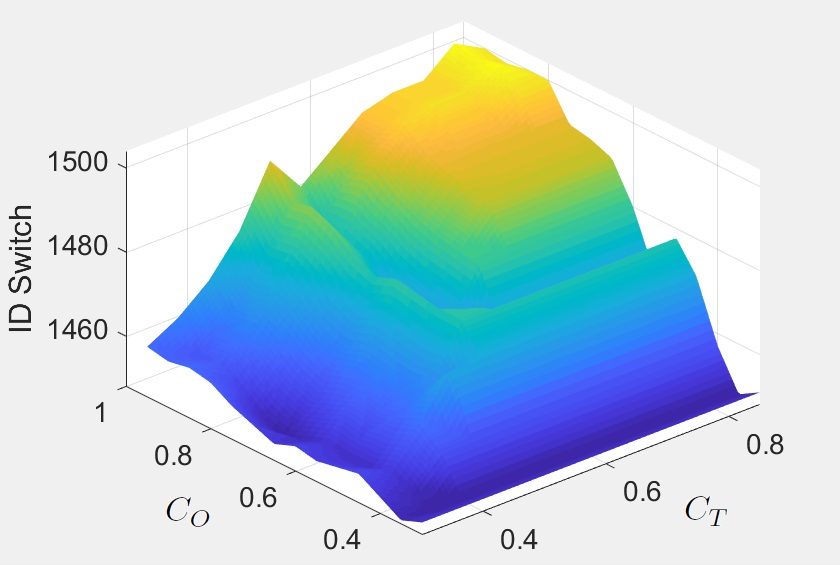}&
\includegraphics[width=5cm, height=3.75cm]{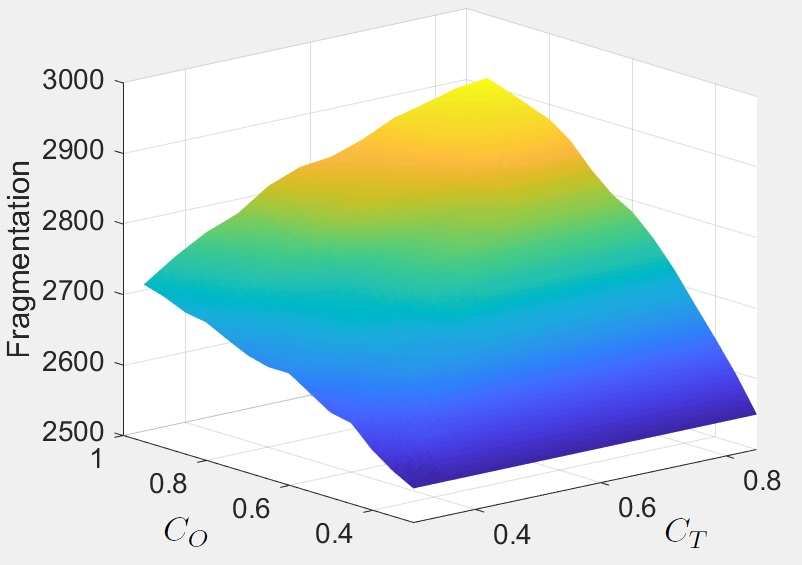}\\
(a)& (b)& (c)\\
\end{tabular}
\caption{The changes of three evaluation metrics on confidence thresholds are presented; (a) MOTA; (b) IDS; (c) FM}
\label{fig:res_3d}
\end{figure*}

\begin{figure*}[htb]
\centering
\includegraphics[width=8.9cm, height=4.2cm]{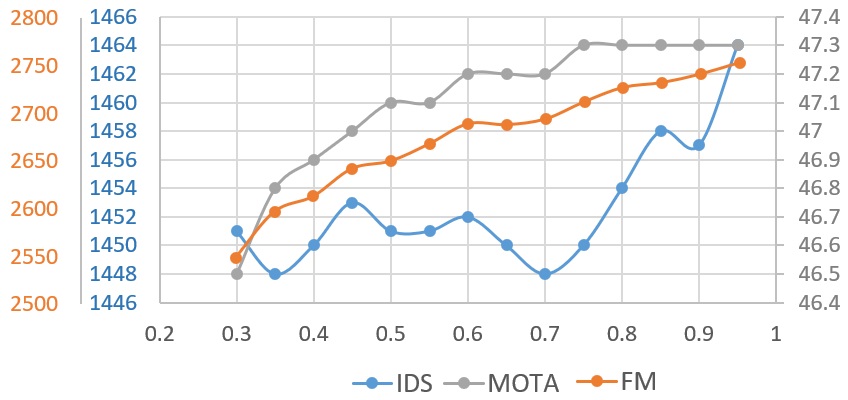}\\
\caption{MOTA, IDS and FM versus $C_O$ for a fixed value of $C_T$}
\label{fig:res_2d}
\end{figure*}

\subsection{Sensitivity to parameters}

The proposed algorithm relies on several parameters from which $C_O$ and $C_T$, i.e. the confidence thresholds for detecting an occluded target, are the most important ones. Thus, the sensitivity of the algorithm has been evaluated with respect to these two parameters. The sensitivity evaluation is done on the train data only since the test data was not accessible for exhaustive analysis. The evaluation is done on both MOT16 and MOT17 datasets, separately. In Fig.~\ref{fig:res_3d}, the changes in MOTA, IDS, and FM metrics based on the changes in $C_O$ and $C_T$ are plotted for MOTA17. Evidently, increasing $C_O$ improves MOTA, and increasing $C_T$ slightly decreases it. Increasing both $C_O$ and $C_T$ increases IDS in general but there are intermediate values in which IDS is near to its minimum. Also, increasing both $C_O$ and $C_T$ increases the FM which is not desired. To optimize the performance of the proposed algorithm, a cost function proportional to the IDS and proportional to the inverse of the MOTA is defined. After minimizing this cost function, the best results for MOT17 are achieved with $C_O=0.75$ and $C_T=0.35$. The general behavior of the algorithm is the same on MOT16 too and $C_O=0.9$ and $C_T=0.55$ are chosen for it. 
Tuning these parameters can indeed improve the performance of the algorithm, but the important point is that the sensitivity of the algorithm to these parameters is low. For example in MOT17, the minimum and maximum of MOTA for the whole range of $C_O$ and $C_T$ is 46.5 and 47.3, respectively which is less than 1 percent. Also the minimum and maximum of IDS are 1448 and 1504 which is less than 4 percent. In addition, for a fixed value of $ C_T$, MOTA, IDS and FM are plotted versus $C_O$ in Fig~\ref{fig:res_2d}. Needles to say, by increasing $C_O$ MOTA and FM grow and IDS changes non-monotonically. To balance between these metrics, 0.75 is chosen for $C_O$ at which MOTA is maximum, IDS is near its minimum and FM is not too high.

\section{Conclusion}
\label{sec:conclusion}

In this paper, a novel algorithm for tracking multiple objects is proposed to handle occlusions and re-identification of lost targets efficiently. This algorithm only uses geometric cues including the location and size of the bounding box of detections. As a result, it is very fast and is appropriate for real-time applications. The performance of this algorithm is comparable to other algorithms that use appearance features and could reach comparable results in terms of MOTA, MOTP, and IDS for the MOT16 dataset in comparison to the Deep SORT algorithm. For the FM metric, the results is better than the Deep SORT algorithm. Moreover, in the MOT17 benchmark, its result is comparable with the state-of-the-art algorithms for the SDP detection set. The results of the proposed algorithm on the MOT17 dataset show that if the detection algorithm has good performance, there is no need to use appearance features to achieve desired performance and only geometric cues can be used to achieve a much higher speed.

\bibliographystyle{IEEEbib}
\bibliography{refs}

\end{document}